\documentclass[times,twocolumn,final]{elsarticle}

\usepackage{url}

\usepackage[latin1,utf8]{inputenc}
\usepackage{times}
\usepackage{epsfig}
\usepackage{graphicx}
\usepackage{amsmath}
\usepackage{amssymb}
\usepackage{color}
\usepackage{caption}
\usepackage{subcaption}     
\usepackage{booktabs}
\usepackage{xspace}

\makeatletter
\DeclareRobustCommand\onedot{\futurelet\@let@token\@onedot}
\def\@onedot{\ifx\@let@token.\else.\null\fi\xspace}

\def\eg{\emph{e.g}\onedot} 
\def\ie{\emph{i.e}\onedot}

\def\etal{\emph{et al}\onedot}
\makeatother

\newcommand{\red}[1]{#1}

\newcommand{\prcite}[1]{~\cite{#1}}
\newcommand{\fakecite}[1]{} %

\begin{document}

\begin{frontmatter}

\title{Rethinking Interactive Image Segmentation: Feature Space Annotation}

\author[unicamp,esiee]{Jord\~ao Bragantini\corref{mycorrespondingauthor}\footnote{Work done during an internship at LIGM.}}
\cortext[mycorrespondingauthor]{Corresponding author}
\ead{jordao.bragantini@gmail.com}

\author[unicamp]{Alexandre X. Falc\~ao}
\ead{afalcao@ic.unicamp.br}

\author[esiee]{Laurent Najman}
\ead{laurent.najman@esiee.fr}

\address[unicamp]{University of Campinas, Laboratory of Image Data Science, Brazil}
\address[esiee]{Université Gustave Eiffel, LIGM, Equipe A3SI, ESIEE, France} 

\begin{abstract}
\begin{sloppypar}
Despite the progress of interactive image segmentation methods, high-quality pixel-level annotation is still time-consuming and laborious --- a bottleneck for several deep learning applications. We take a step back to propose interactive and simultaneous segment annotation from multiple images guided by feature space projection.
This strategy is in stark contrast to existing interactive segmentation methodologies, which perform annotation in the image domain. 
We show that feature space annotation achieves competitive results with state-of-the-art methods in foreground segmentation datasets: iCoSeg, DAVIS, and Rooftop.  Moreover, in the semantic segmentation context, it achieves 91.5\% accuracy in the Cityscapes dataset, being 74.75 times faster than the original annotation procedure. Further, our contribution sheds light on a novel direction for interactive image annotation that can be integrated with existing methodologies.
The supplementary material presents video demonstrations. Code available at \url{https://github.com/LIDS-UNICAMP/rethinking-interactive-image-segmentation}.
\end{sloppypar}

\end{abstract}


\end{frontmatter}

\section{Introduction}

Convolutional Neural Networks (CNNs) can achieve excellent results on image classification\red{~\cite{He:2016:Resnet}}, image segmentation\red{~\cite{Wang:2020:HRNet}}, pose detection\red{~\cite{Sun:2019:HRPose}}, and other images/video-related tasks\red{~\cite{Wu:2020:VideoAttention}}, at the cost of an enormous amount of high-quality annotated data and processing power. Thus, interactive image segmentation with reduced user effort is of primary interest to create such datasets for the training of CNNs. Concerning image segmentation tasks, the annotations are pixel-wise labels, usually defined by interactive image segmentation methods \fakecite{Couprie:2011:PowerWatershed, Falcao:2004:IFT, Grady:2006:RandomWalks,McGuinness:2010:ComparativeInteractive}\prcite{McGuinness:2010:ComparativeInteractive} or by specifying polygons in the object boundaries~\cite{Russell:2008:LabelMe}.

\begin{figure}[!ht]
    \centering
    \includegraphics[width=0.75\linewidth]{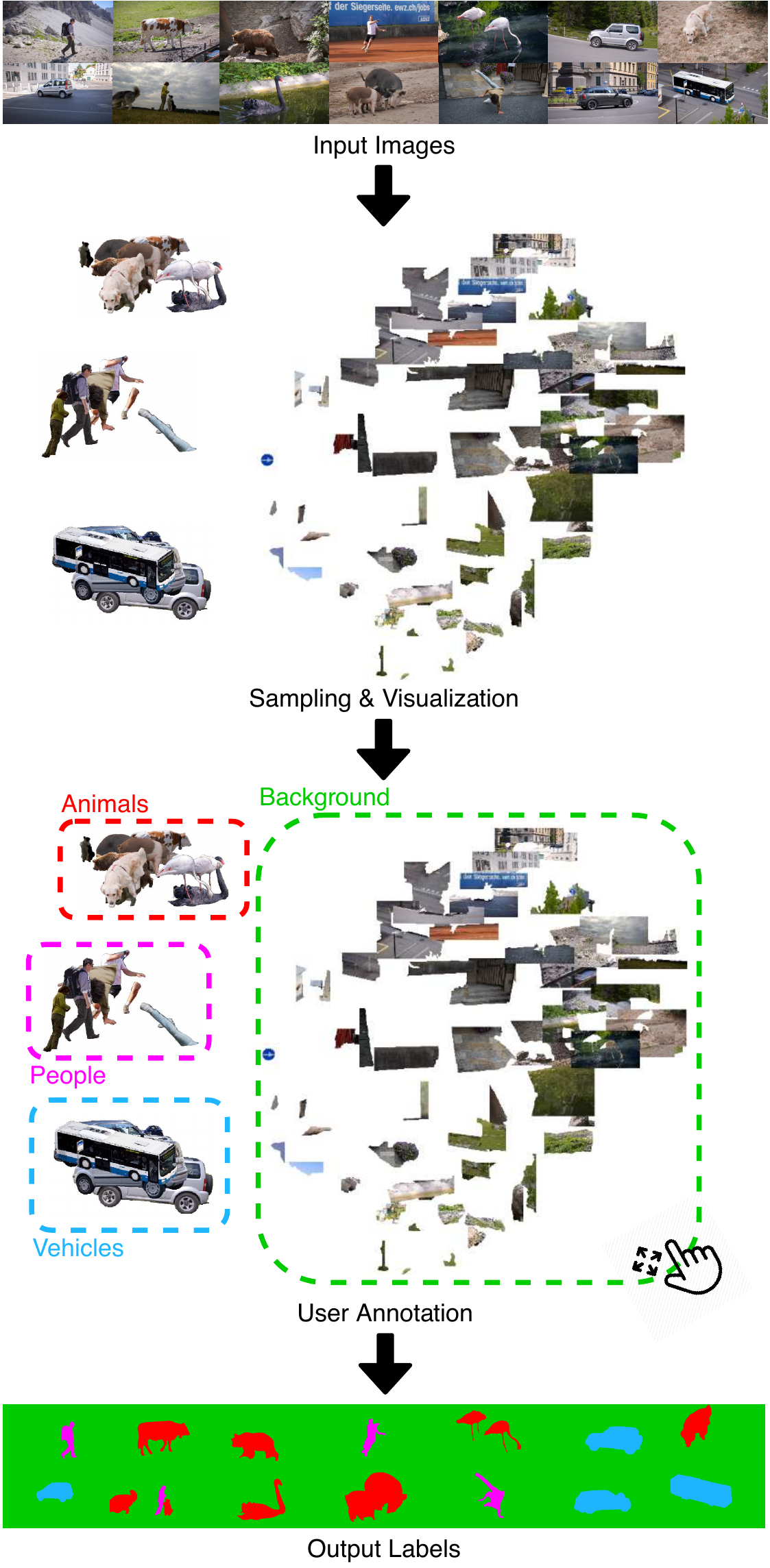}
    \caption{Our approach to interactive image segmentation: candidate segments are sampled from the dataset and presented in groups of similar examples to the user, who annotates multiple segments in a single interaction.}
    \label{fig:abstract}
\end{figure}

Recent interactive image segmentation methods based on deep learning can significantly reduce user effort by performing object delineation from a few clicks, sometimes in a single user interaction\fakecite{Li:2018:LatentDiversity, Lin:2020:FirstClickAtte, Zhang:2020:IOG}\prcite{Lin:2020:FirstClickAtte, Zhang:2020:IOG}. However, such deep neural networks do not take user input as hard constraints and so cannot provide enough user control. Novel methods can circumvent this issue by refining the neural network's weights while enforcing the correct results on the annotated pixels\fakecite{Jang:2019:BRS, Sofiiuk:2020:fBRS, Kontogianni:2020:ContinuousAdapt}\prcite{Sofiiuk:2020:fBRS,Kontogianni:2020:ContinuousAdapt}. Their results are remarkable for foreground segmentation. Still, in complex cases or objects unseen during training, the segmentation may be unsatisfactory even by extensive user effort.

The big picture in today's image annotation tasks is that thousands of images with multiple objects require user interaction. While they might not share the same visual appearance, their semantics are most likely related. Hence, thousands of clicks to obtain thousands of segments with similar contexts do not sound as appealing as before.

This work presents a scheme for interactive large-scale image annotation that allows user labeling many similar segments at once. It starts by defining segments from multiple images and computing their features with a neural network pre-trained from another domain. User annotation is based on feature space projection, Figure~\ref{fig:abstract}. As it progresses, the similarities between segments are updated with metric learning, increasing the discrimination among classes, and further reducing the labeling burden.

\red{Our implementation of this methodology is represented in Figure~\ref{fig:pipeline} and it is described with further details in Section~\ref{sec:method}.}

\textbf{Contribution:} To our knowledge, this is the first interactive image segmentation methodology that does not receive user input on the image domain. Hence, our goal is not to beat the state-of-art of interactive image segmentation but to demonstrate that other forms of human-machine interaction, notably feature space interaction, can benefit the interactive image segmentation paradigm and can be combined with existing methods to perform more efficient annotation.

\section{Related Works}
\label{sec:relworks}

\red{This section is divided into two parts, a first section where methodologies related to pixel labeling are presented and a second section where we review auxiliary techniques that are employed in the proposed pipeline but are not interactive segmentation methods.}

\subsection{\red{Interactive segmentation and data annotation}}

In this work, we address the problem of assigning a label (\ie class) to every pixel in a collection of images, denoted as image annotation in the remaining of the paper. This is related to the foreground (\ie region) segmentation microtasks, where the region of interest has no specific class assigned to it, and the delineation of the object is of primary interest --- in standard image annotation procedures, this is the step preceding label assignment~\cite{Russell:2008:LabelMe}. 

Current deep interactive segmentation methods, from click\fakecite{Xu:2016:DeepObjSel,Li:2018:LatentDiversity, Maninis:2018:DEXTR,Kontogianni:2020:ContinuousAdapt, Jang:2019:BRS,Sofiiuk:2020:fBRS, Lin:2020:FirstClickAtte}\prcite{Sofiiuk:2020:fBRS, Lin:2020:FirstClickAtte,Kontogianni:2020:ContinuousAdapt}, bounding-box\fakecite{Maninis:2018:DEXTR, Benenson:2019:LargeScale, Zhang:2020:IOG}\prcite{Zhang:2020:IOG} to polygon-based approaches\fakecite{Castrejon:2017:PolyRNN,Acuna:2018:PolyRNN++,Ling:2019:CurveGCN}\prcite{Ling:2019:CurveGCN} address this microtask of segmenting and then labeling each region individually to generate annotated data.

A minority of methods segment multiple objects jointly; to our knowledge, in deep learning, this has been employed only once~\cite{Agustsson:2019:Jointly}; in classical methods, a hand-full could do this efficiently\fakecite{Couprie:2011:PowerWatershed, Falcao:2004:IFT, Grady:2006:RandomWalks}\prcite{Falcao:2004:IFT}.

Fluid annotation~\cite{Andriluka:2018:FluidAnnot} proposes a unified human-machine interface to perform the complete image annotation; the user annotation process starts from the predictions of an existing model, requiring user interaction only where the model lacks accuracy, further reducing the annotation effort. The user decides which action it will perform at any moment without employing active learning (AL). Hence, the assumption is that the user will take actions that will decrease the annotation budget the most.
    
This approach falls in the Visual Interactive Labeling (VIAL)~\cite{Bernard:2018:VIAL} framework, where the user interface should empower the users, allowing them to decide the optimal move to perform the task efficiently. Extensive experiments~\cite{Bernard:2017:VAvsAL} have shown that this paradigm is as competitive as AL and obtains superior performance when starting with a small amount of annotated data. 

Inside the VIAL paradigm, feature space projection has been employed for user guidance in semi-supervised label propagation~\cite{Benato:2018:SemiSupLearn, Benato:2019:SemiProjAnnot} and for object detection in remote-sensing~\cite{Vargas:2019:CoconutAnnot}. However, its use for image segmentation has not been explored yet.

\subsection{\red{Complementary background}}
\label{subsec:comp}

\red{In this section, we briefly review necessary concepts to the proposed pipeline, them being: boundary prediction and its relationship to image segmentation; and metric learning and dimensionality reduction.}

\paragraph{\red{Boundary prediction (\ie edge detection)}} \red{These methods regress pixels intensities of the transitions between homogeneous regions (\ie boundary), where a greater value indicates a stronger boundary and a lower value a weaker. The notion of weaker and stronger is context dependent, usually being related to how discrepant regions are from their neighbors.}

\red{The duality between boundaries (\ie contour) and segmentation is such a binary boundary, where background is zero and the contour is one, contains a segment delineated by the contour; on the fuzzy scenario where the contours' values are between zero and one, multiple sets of segments can be obtained by creating a binary contour consisting of only the pixels with values above a threshold in the fuzzy contour, and computing segmentation as described in the binary case. Moreover, an increasing sequence of thresholdings produce a decreasing set of disjoint segments, such that the previous segments are a subset of the subsequent thresholdings, producing a hierarchical segmentation. This fuzzy contour representation is known as a ultrametric contour map --- an interested reader can refer to~\cite{Najman:1996:Watershed, Najman:2011:EquivalenceWatershed} to review the duality between contours and hierarchies.}

\red{In Convolutional Oriented  Boundaries~\cite{Maninis:2017:COB}, a CNN predicts multiple boundaries in multiple scales and orientations and combine them into an ultrametric contour map (\ie fuzzy contour) to perform the hierarchical segmentation}.

\red{Holistically-nested Edge Detection (HED)~\cite{Xie:2015:Holistically} employed a CNN to predict boundaries at multiple scales in what they called side predictions, starting from the shallower layers of the network up to the last layer, and fusing them into a single output image. Their loss function optimizes each side prediction independently and the combined final output. Lie~\etal~\cite{Liu:2017:RCFEdgeDetec} improved upon HED by learning to fuse side outputs using $1 \times 1$ convolutional blocks. Hybrid Convolutional Features~\cite{Hu:2018:LearningHybrid} learns an additional parameter to normalize the features of different side predictions before the multi-scale fusion, balancing the influence of earlier and latter feature blocks. Liu~\etal~\cite{Liu:2021:SemanticEdgeDetection} incorporates semantic labels by predicting two different outputs, a class-agnostic edge detection, as the other approaches, and a edge-detection with class predictions. Further boosting the edge prediction performance by using higher-level semantic information.}

\red{Other tasks also employ edge estimation to enhance their performance, notably PoolNet~\cite{Liu:2019:PoolNet} switches between saliency object prediction and edge estimation in the training loop with the same architecture to obtain saliency with greater boundary adherence.}

\paragraph{\red{Dimensionality reduction and metric learning}} 

\red{Both of these techniques concerns with learning a transformation (\ie function) or an embedding to aid a task of interest. The former being in the unsupervised scenario and in the recent literature, mostly embedding into a 2-dimensional space for data visualization and exploratory analysis; and the latter, being supervised or semi-supervised.}

\red{In this work, the dimensionality reduction is used to arrange the data on a 2D plane for visualization and the metric learning to update their positions as the user annotates, facilitating subsequent annotations.}

\red{Dimensionality reduction aims at reducing a feature space from a higher to a lower dimension with similar characteristics. Some methods enforce the global structure (\eg PCA); other approaches, such as non-linear methods, focus on local consistency, penalizing neighborhood disagreement between the higher and lower dimensional spaces.}

\red{In some applications, the dimensionality reduction aims to preserve the original features' characteristics. In our case, we wish to facilitate the annotation as much as possible, hence, a reduction that groups similar segments and segregates dissonant examples is more beneficial than preserving the original information.}

\red{The t-SNE~\cite{Maaten:2008:tSNE} algorithm is the most used technique for non-linear dimensionality reduction. It projects the data into a lower-dimensional space while minimizing the divergence between the higher- and lower-dimensional neighborhood distributions.}

\red{Uniform Manifold Approximation (UMAP)~\cite{McInnes2018:Umap} improves upon t-SNE by providing a geometric and topological theoretical for dimensionality reduction, their implementation produces consistent embedding given multiple executions by using the spectral embedding as initialization, and their optimization process to compute the embedding is faster by not requiring to recompute a score for every pair after each iteration, thus being much faster than t-SNE.}

\red{Initially, metric learning methods were concerned with finding a metric where some distance-based (or similarity) classification~\cite{Weinberger:2009:LMNN, Goldberger:2004:NCA} and clustering~\cite{Xing:2003:DistanceML} would be optimal, in the sense that samples from the same class should be closer together than adversary examples. Given some regularity conditions, learning this new metric is equivalent to embedding the data into a new space.}

\red{The metric learning objective functions can be roughly divided into two main varieties, soft assignment and triplet-based techniques. The former, as proposed in NCA~\cite{Goldberger:2004:NCA}, maximizes a soft-neighborhood assignment computed through the soft-max function over the negative distance between the data points, penalizing label disagreement of immediate neighbors more than samples further apart. Triplet-based methods~\cite{Weinberger:2009:LMNN} select two examples from the same class and minimize their distance while pushing away a third one from a different class when it violates a threshold given the pair distance. Thus, avoiding unnecessary changes when a neighborhood belongs to a single class.}

\red{More recently, these methods began to focus mostly on improving embedding through neural networks rather than on the metric-centered approach.}

\section{Proposed Method}
\label{sec:method}

Our methodology allows the user to annotate multiple classes jointly and requires lower effort when the data are redundant by allowing multiple images to be annotated at once. Multiple image annotation is done by extracting candidate regions (\ie segments) and presenting them simultaneously to the user. The candidate segments are displayed closer together according to their visual similarity~\cite{Sacha:2016:VisualInterDR} for the label assignment of multiple segments at once. 
Hence, user interaction in the image domain is only employed when necessary --- not to assign labels but to fix incorrect segments. Since, this action is the one that requires the most user effort and has been the target of weakly-supervised methodologies\fakecite{Araslanov:2020:SingleStage, Zhou:2018:PRM, Zhu:2019:InstPRM, Hsu:2019:DeepCO3}\prcite{Zhu:2019:InstPRM} that try to avoid it altogether. Therefore, this approach is based on the following pillars:
\begin{itemize}
	\item The segment annotation problem should be evaluated as a single task~\cite{Andriluka:2018:FluidAnnot}. While dividing the problem into microtasks is useful to facilitate the user and machine interaction, they should not be treated independently since the final goal is the complete image annotation. 
	\item The human is the protagonist in the process, as described in the VIAL process~\cite{Bernard:2018:VIAL}, deciding which action minimizes user effort for image annotation while the machine assists in well-defined microtasks.
	\item The annotation in the image domain is burdensome; thus, it should be avoided, but not neglected, since perfect segmentation is still an unrealistic assumption.
	\item The machine should assist the user initially, even when no annotated examples are present~\cite{Bernard:2017:VAvsAL}, and as the annotation progresses, labeling should get easier because more information is provided.
\end{itemize}

\subsection{Overview}

\begin{figure}[!ht]
    \centering
    \includegraphics[width=0.85\columnwidth]{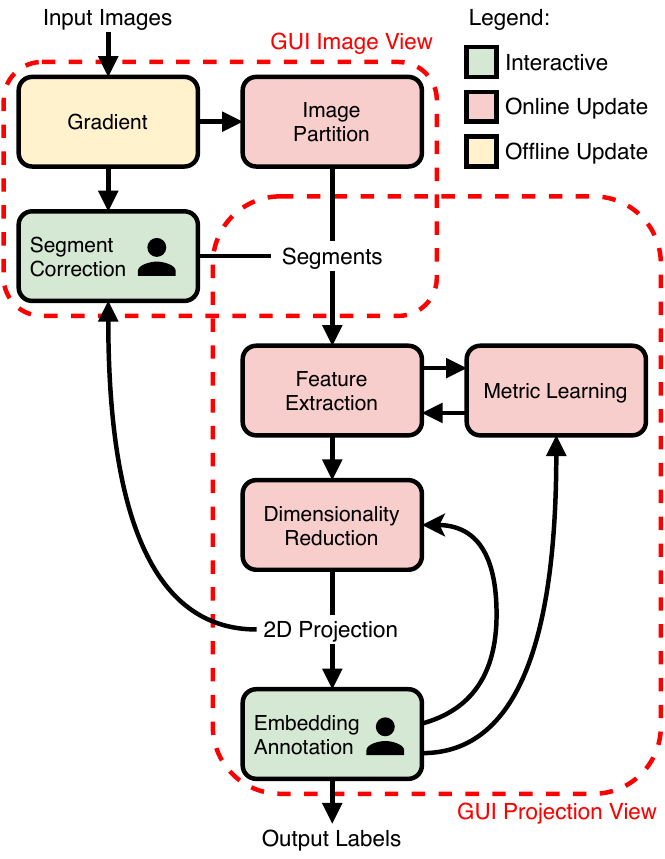} 
    \caption{The proposed feature space annotation pipeline. \red{Input images' gradients (\ie edge detection) are estimated using a neural network, from these gradients we partition the images into a set of candidate segments. Another network extracts the segments' features, which are used to compute their 2D coordinates (\ie embedding), the segments are arranged at their respective coordinates and shown to the user for annotation. As labels are provided, the segments' features and coordinates are updated using metric learning to cluster samples belonging to the same class. The user can also select segments on the embedding for correction on the image domain. Upon conclusion of the annotation, the assigned labels are mapped to the original image domain using the segments' regions.}}
    \label{fig:pipeline}
\end{figure}

The proposed methodology is summarized in Figure~\ref{fig:pipeline} and each component is described in the subsequent sections.

The user interface is composed of two primary components, the Projection View and the Image View. Red contours in Figure~\ref{fig:pipeline} delineate which functionalities are present in these widgets. The Projection View is concerned with displaying the segments arranged in a canvas (Figure~\ref{fig:abstract}), enabling the user to interact with it: assigning labels to clusters, focusing on cluttered regions, and selecting samples for segmentation inspection and correction in the image domain.

Image View displays the image containing a segment selected in the canvas. It is highlighted to allow fast component recognition among the other segments' contours. Samples already labeled are colored by class. This widget allows further user interaction to fix erroneous delineation, as further discussed in Section~\ref{subsec:correct}. Segment selection also works backwardly, when a region is selected in the Image View, its feature space neighborhood is focused on the projection canvas, accelerating the search for relevant clusters by providing a mapping from the image domain to the projection canvas.

The colored rectangles in Figure~\ref{fig:pipeline} represent data processing stages: yellow represents fixed operations that are not updated during user interaction, red elements are updated as the user annotation progresses, and the greens are the user interaction modules. Arrows show how the data flows in the pipeline.

The pipeline works as follows, starting from a collection of images, their gradients are computed to partition each image into segments, which will be the units processed and annotated in the next stages.

Since we wish to cluster together similar segments, we must define a similarity criterion. Therefore, for each segment, we obtain their deep representation (\ie features). Their Euclidean distances are used to express this information --- they are more dissimilar as they are further apart in the feature space.

The next step concerns the notion of similarity between segments as presented and perceived by the user. We propose communicating this information to the user by displaying samples with similar examples in the same neighborhood. Hence, the segments' features are used to project them into the 2D plane while preserving, as best as possible, their relative feature space distances.

The user labeling process is executed in the 2D canvas by defining a bounding-box and assigning the selected label to the segments inside it. As the labeling progresses, their deep representation is updated using metric learning, improving class separability, enhancing the 2D embedding, thus, reducing the annotation effort. We refer to~\cite{Bernard:2018:VIAL} for a review in visual interactive labeling and~\cite{Sacha:2016:VisualInterDR} for interactive dimensionality reduction systems.

This pipeline relies only upon the assumption that it is possible to find meaningful candidate segments from a set of images and extract discriminant features from them to cluster together similar segments. Even though these problems are not solved yet, existing methods can satisfy these requirements, as they are validated in our study of parts (Section~\ref{subsec:ablation}).

\subsection{Gradient and Image Partition}
\label{subsec:impart}

Gradient computation and watershed-based image partition are operations of the first step of the pipeline, obtaining initial candidate segments. In the ideal scenario, the desired regions are represented by a single connected component, requiring no further user interaction besides labeling.

However, obtaining meaningful regions is a challenging and unsolved problem. Desired segments vary from application to application. On some occasions, users wish to segment humans and vehicles in a scene, while in the same image, other users may desire to segment the clothes and billboards. Thus, the proposed approach has to be class agnostic and enables the user to obtain different segment categories without effort.

The usual approach computes several solutions (\eg Multiscale Combinatorial Grouping (MCG)~\cite{Pont:2016:MCG}) and employs a selection policy to obtain the desired segments. However, it does not guarantee disjoint regions, and it generates thousands of candidates, further complicating the annotation process.

Given that segments should be disjoint, and they are also task dependent, we chose to employ hierarchical segmentation techniques for this stage. Most of the relevant literature for this problem aims at obtaining the best gradient (edge saliency) to compute segmentation, \red{as described in the boundary prediction paragraph of Section~\ref{subsec:comp}.}

We opted to employ the flexible hierarchical watershed framework~\cite{Cousty:2018:Hierarchical, Cousty:2011:IncrAlgo} for delineating candidate segments on the gradient image estimated by PoolNet~\cite{Liu:2019:PoolNet} architecture\red{, because we noticed that this approach produces less irrelevant boundaries for image segmentation than other methods.}
The hierarchical watershed allows manipulation of the region merging criterion, granting the ability to rapidly update the segments' delineation. Besides, hierarchical segmentation lets the user update segments without much effort (\eg, obtaining a more refined segmentation by reducing the threshold, but as a trade-off, the number of components increases). Further, a watershed algorithm can also interactively correct the delineated segments (Section~\ref{subsec:correct}).

Starting from a group of $N$ images without annotations, $\{I_1, I_2,\ldots, I_N\}$, their gradient images are computed. For each gradient $G_i$, $i\in [1,N]$, its watershed hierarchy is built and disjoint segments $\{S_{i,1},\ldots,S_{i,n_i}\}$ are obtained by thresholding the hierarchy. The required parameters (threshold and hierarchy criterion) are robust and easy to be defined by visual inspection on a few images. More details about that are presented in Section~\ref{subsec:correct}.

\subsection{Feature Extraction}
\label{subsec:feats}

Before presenting the regions arranged by similarity, a feature space representation where dissonant samples are separated must be computed. For that, we refer to CNN architectures for image classification tasks, without their fully connected layers\fakecite{Long:2015:FCNN} used for image classification.

Each segment is treated individually; we crop a rectangle around the segment in the original image, considering an additional border to not impair the network's receptive field. In this rectangle, pixels that do not belong to the segment are zeroed out. Otherwise, segments belonging to the same image would present similar representations. The segment images are then resized to $224 \times 224$ and forwarded through the network, which outputs a high-dimensional representation, $\phi_{i,j}$. In this instance $\phi_{i,j} \in \mathbb{R}^{2048}$, for each $S_{i,j}$. We noticed that processing the segment images without resizing them did not produce significant benefits and restricted the use of large batches' efficient inference.

Since our focus is on image annotation, where labeled data might not be readily available, feature extraction starts without fine-tuning. It is only optimized as the labeling progresses. Any CNN architecture can be employed, but performance is crucial. We use the High-Resolution Network (HRNet)~\cite{Wang:2020:HRNet} architecture, pre-trained on ImageNet without the fully connected layers. It is publicly available with multiple depths, and its performance is superior to other established works for image classification, such as ResNet~\cite{He:2016:Resnet}. During the development of this work, other architectures were proposed\fakecite{Tan:2019:Efficientnet, Radosavovic:2020:RegNet}\prcite{Radosavovic:2020:RegNet}, significantly improving the classification performance while using comparable computing resources. They were not employed in our experimental setup, but it might improve our results.

\subsection{Dimensionality Reduction}
\label{subsec:proj}

\red{For dimensionality reduction we employed the previously mentioned UMAP~\cite{McInnes2018:Umap} (Section~\ref{subsec:comp})}, for the following reasons: the projection is computed faster, samples can be added without fitting the whole data, its parameters seem to provide more flexibility to choose the projection scattering --- enforcing local or global coherence --- and most importantly, it allows using labeled data to enforce consistency between samples of the same class while still allowing unlabeled data to be inserted.

Note that dimensionality reduction is critical to the whole pipeline because it arranges the data to be presented to the user, where most of the interaction will occur.

\begin{figure}[ht]
    \centering
    \includegraphics[width=0.95\linewidth]{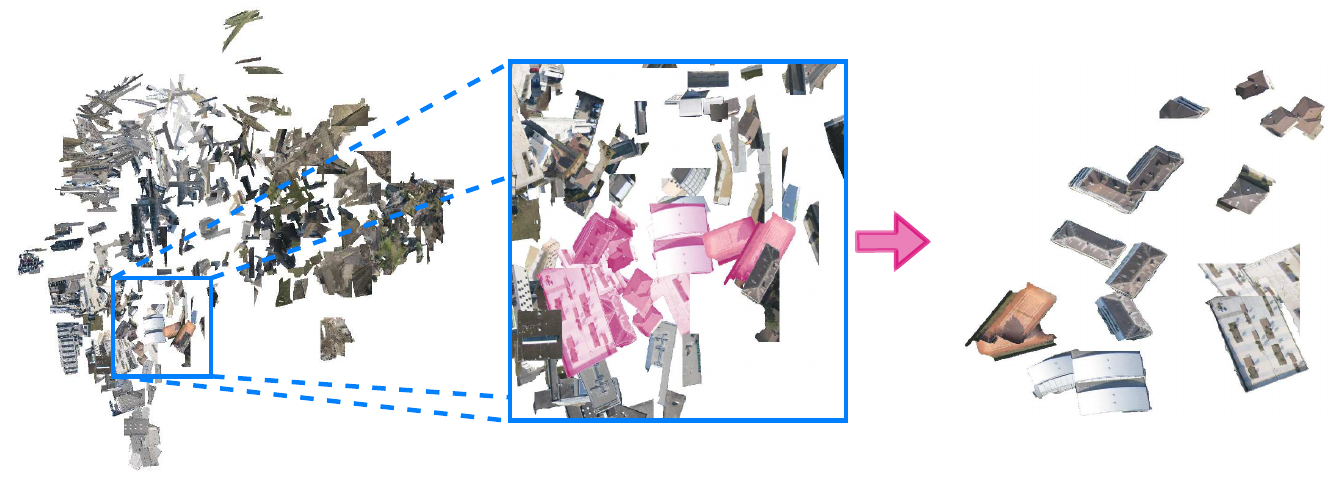} 
    \caption{Local re-projection example: Global projection with a region highlighted in blue; A subset of segments is selected by the user, in pink; Their local embedding is computed for a simpler annotation.}
    \label{fig:zoomin}
\end{figure}

The 2D embedding can produce artifacts, displaying distinct segments clustered together due to the trade-off between global and local consistency even though they might be distant in the higher-dimensional feature space. Therefore, the user can select a subset of samples and interact with their local projection in a pop-up window, where the projection parameter is tuned to enforce local consistency. The locally preserving embedding (Figure~\ref{fig:zoomin}) separates the selected cluttered segments (in pink) into groups of similar objects (tennis court, big households, small households, etc.), making it easier for label assignment.

On images, CNN's features obtain remarkable results consistent with the human notion of similarity between objects. Considering that an annotator evaluates images visually, sample projection is our preferred approach to inform the user about possible clusters, as explained in the next section. Other visualizations must be explored for other kinds of data, such as sound or text, where the user would have difficulty to visually exploit the notion of similarity~\cite{Bernard:2018:VIAL, Sacha:2016:VisualInterDR}.

\subsection{Embedding Annotation}
\label{subsec:annot}

Each segment is displayed on their 2-dimensional coordinate, as described in the previous section. To annotate a set of segments, the user selects a bounding-box around them in the canvas, assigning the designed label. Hence, each $S_{i,j}$ inside the defined box is assigned to a label $L_{i,j}$. Finally, to obtain annotated masks in the image domain, the label $L_{i,j}$ of a segment $S_{i,j}$ is mapped to its pixels in $I_i$, thus resulting in an image segmentation (pixel annotation).

Due to several reasons, such as spurious segments or over-segmented objects, a region could be indistinguishable. Hence, when a single segment is selected in the projection, its image is displayed in the Image View as presented earlier. This action also works backwardly, the user can navigate over the images, visualize the current segments, and upon selection, the segments are focused in the projection view. Thus, avoiding the effort of searching individual samples in the segment scattering.

Additional care is necessary when presenting a large number of images, mainly if each one contains several objects, because the number of segments displayed on the canvas may impair the user's ability to distinguish their respective classes for annotation. Therefore, only a subset of the data is shown to the user initially. Additional batches are provided as requested while the labeling and the embedding progress, reducing the annotation burden.

\subsection{Metric Learning}
\label{subsec:metric}

The proposed pipeline is not specific to any objects' class and does not require pre-training, but as the annotation progresses, the available labels can be employed to reduce user effort --- less effort is necessary when the clusters are homogeneous and not spread apart.

\begin{figure}
    \centering
    \begin{subfigure}[b]{0.45\columnwidth}
        \centering
        \fbox{\includegraphics[width=\textwidth]{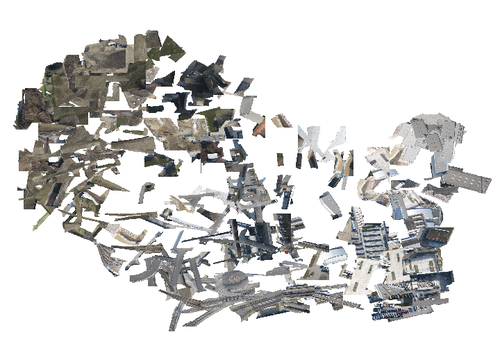}}
        \caption{}
    \end{subfigure}
    \hfill
    \begin{subfigure}[b]{0.45\columnwidth}
        \centering
        \fbox{\includegraphics[width=\textwidth]{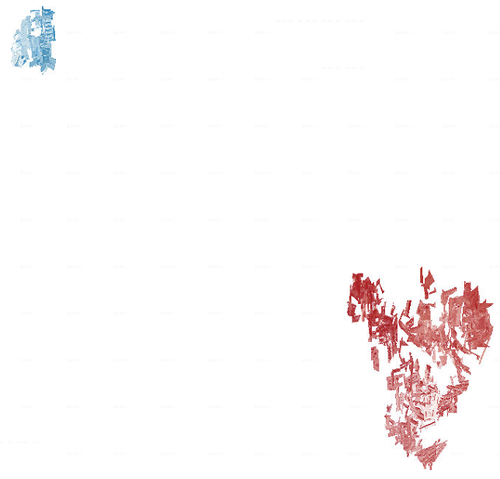}}
        \caption{}
    \end{subfigure} \\
    \begin{subfigure}[b]{\columnwidth}
        \centering
        \fbox{\includegraphics[width=\textwidth]{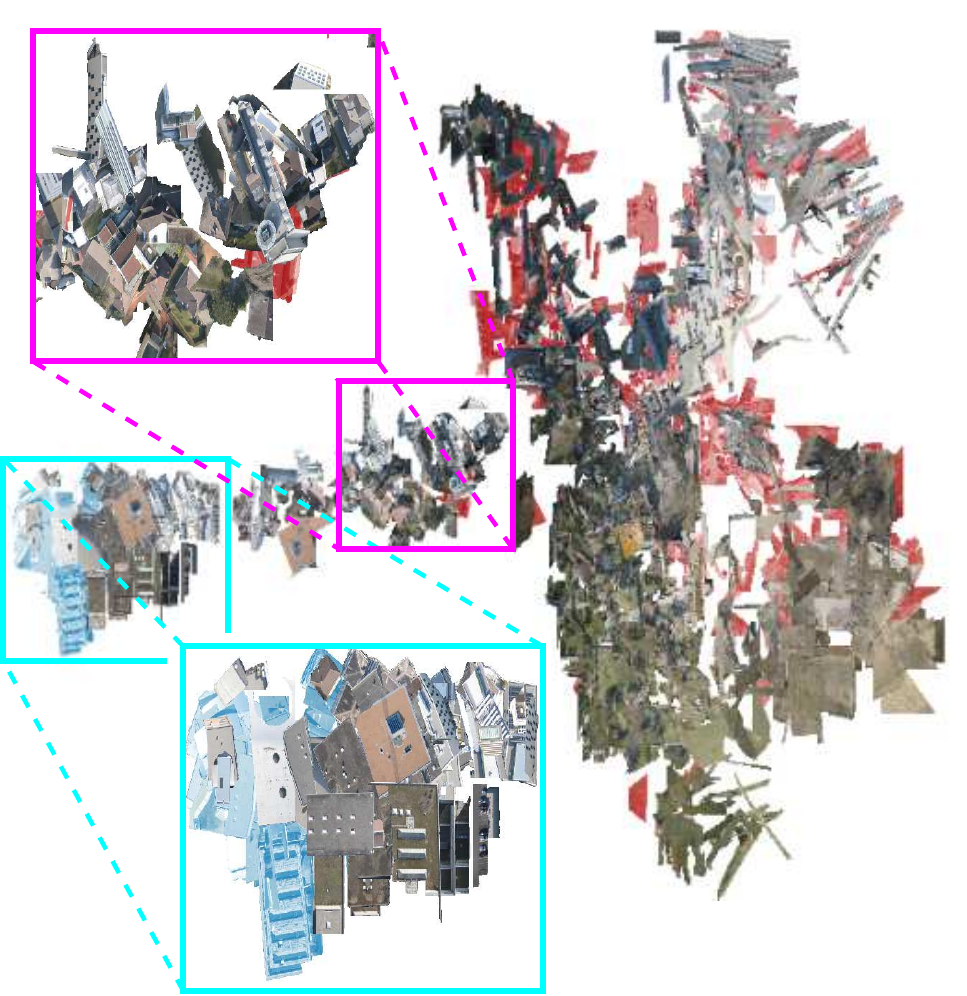}}
        \caption{}
    \end{subfigure}
    \caption{Example of metric learning in the Rooftop dataset: (a) Segment arrangement from an initial batch (10 images). (b) Displacement after labeling and employing metric learning, foreground (rooftops) in blue and background in red. (c) Projection with additional unlabeled data (plus 20 images). Most rooftops' segments are clustered together (the cyan box).  The magenta box indicates clusters from mixed classes and spurious segments. They suggest where labels are required the most for a next iteration of labeling and metric learning. The remaining clusters can be easily annotated.}
    \label{fig:metriclearn}
\end{figure}

For that, we employ a \emph{metric learning} algorithm. Figure~\ref{fig:metriclearn} shows an example of how metric learning can make clusters of a same class more compact and better separate clusters from distinct classes in our application. 

In our pipeline, the original large-margin loss~\cite{Weinberger:2009:LMNN} was employed, using our previously mentioned feature extractor network, due to its excellent performance with only a single additional parameter. We follow here Musgrave \etal~\cite{Musgrave:2020:MLRealityCheck}, which showed that some novel methods are prone to overfitting and require more laborious parameter tuning. 

\subsection{Segment Correction}
\label{subsec:correct}

Since segment delineation is not guaranteed to be perfect, component correction is crucial, especially for producing ground-truth data, where pixel-level accuracy is of uttermost importance. Hence, segments containing multiple objects (under segmentation) are corrected by positive and negative clicks, splitting the segment into two new regions according to the user's positive and negative cues.

Here we use a classical graph-based algorithm as we focus mainly on the feature space annotation; more modern CNN-based approaches can be employed on a real scenario.

Given an under-segmented region $S_{i,j}$, we define an undirected graph $\mathcal G = (V, E, w)$ where the vertices $V$ are the pixels in $S_{i,j}$, the edges connect 4-neighbors, constrained to be inside $S_{i,j}$, and each edge $(p,q)\in E$ is weighted by $w(p, q) = (G_{i,j}(p) + G_{i,j}(q))/2$. A segment partition is obtained by the image foresting transform~\cite{Falcao:2004:IFT} algorithm for the labeled watershed operator given two sets of clicks $C_{pos}$ and $C_{neg}$ as defined by the user. This operation offers full control over segmentation, is fast, and improves segment delineation.

Further, the user can change the hierarchical criterion for watershed segmentation, preventing interactive segment correction in multiple images. For example, in an image overcrowded with irrelevant small objects, the user can bias the hierarchy to partition larger objects by defining the hierarchy ordering according to the objects' area in the image domain, filtering out the spurious segments.

Therefore, the Image View interface allows inspection of multiple hierarchical segmentation criteria and their result for given a threshold. The segments are recomputed upon user confirmation, maintaining the labels of unchanged segments. Novel segments go through the pipeline for feature extraction and projection into the canvas.

\section{Experiments}
\label{sec:exp}

This section starts by describing the datasets chosen for the experiments and the implementation details for our approach. Since our method partitions the images into segments and solves the simultaneous annotation of multiple objects by interactive labeling of similar segments' clusters, we present a study of two ideal scenarios to evaluate each main step individually. In the first experiment, the images are partitioned into perfect segments constrained to each object's mask, the rest of the pipeline is executed as proposed, evaluating the feature space annotation efficiency. Next, we assess the image partition by assuming optimal labeling, thus, only investigating the initial segmentation performance. Hence, the study of parts evaluates the limitations of the initial unsupervised hierarchical segmentation techniques, description with metric learning, and projection. Subsequently, we compare with state-of-the-art methods and present the quantitative and qualitative results. The supplementary materials include videos of the UI usage and annotation experiments.

Note that our goal goes beyond showing that the proposed method can outperform others. We are pointing a research direction that exploits new ways of human-machine interaction for more effective data annotation.

Typically, a robot user executes the deep interactive segmentation experiments; in contrast, our study is conducted by a volunteer with experience in interactive image segmentation. Thus, we are taking into account the effort required to locate and identify objects of interest.

\subsection{Datasets}

\red{Since the proposed method is concerned with domain-specific annotation and not the micro-task of segmenting a single object, we selected datasets from video segmentation, co-segmentation, and semantic segmentation tasks, in which the objects of interest are to some extent related. In the foreground versus background scenario (items 1,2 and 3), our approach is quantitatively compared to existing foreground segmentation methods, Section~\ref{subsec:quant}. In the semantic segmentation case (item 4), we do qualitative analysis and compare to existing results, Section~\ref{subsec:qual}. The datasets used were:}

\begin{enumerate}
    \item \emph{CMU-Cornell iCoSeg}~\cite{Batra:2011:iCoSeg}: It contains 643 natural images divided into 38 groups.
    Within a group, the images have the same foreground and background but are seen from different point-of-views.
    \item \emph{DAVIS}~\cite{Perazzi:2016:DAVIS}: It is a video segmentation dataset containing 50 different sequences. Following the same procedure as in ~\cite{Jang:2019:BRS}, multiple objects in each frame were treated as a single one, and the same subset of 345 frames (10 \% of the total) was employed.
    \item \emph{Rooftop}~\cite{Sun:2014:Rooftop}: It is a remote sensing dataset with 63 images,
    and in total containing 390 instances of disjoint rooftop polygons.
    \item \emph{Cityscapes}~\cite{Cordts:2016:Cityscapes}: It is a semantic segmentation dataset for autonomous driving research. It contains video frames from 27 cities divided into 2975 images for training, 500 for validation, and 1525 for testing. The dataset  contains 30 classes (\eg roads, cars, trucks, poles), we evaluated using only the 19 default classes.
\end{enumerate}

\subsection{Implementation details}

We implemented a user interface in Qt for Python. To segment images into components, we used Higra\fakecite{Perret:2019:Higra, Code:2020:Higra}\prcite{Perret:2019:Higra} and the image gradients generated with PoolNet~\fakecite{Code:2020:PoolNet}. We computed gradients over four scales, 0.5, 1, 1.5, and 2, and averaged their output to obtain a final gradient image. For the remaining operations, including the baselines, we used NumPy\fakecite{Harris:2020:NumPy}, the PyTorch Metric Learning package~\fakecite{Musgrave:2020:PytorchML} and the available implementations in PyTorch~\fakecite{Paszke:2019:Pytorch}.

For segment description with metric learning, we used the publicly available \emph{HRNet-W18-C-Small-v1}~\fakecite{Code:2020:HRNet} configuration pre-trained on the ImageNet dataset. In the metric-learning stage, the Triplet-Loss margin is fixed at 0.05. At each call, the embedding is optimized through Stochastic Gradient Descent (SGD) with momentum of 0.8 and weight decay of 0.0005 over three epochs with 1000 triplets each. The learning rate starts at 0.1 and, at each epoch, it is divided by 10.

We used UMAP~\cite{McInnes2018:Umap} with 15 neighbors for feature projection and a minimum distance of 0.01 in the main canvas. The zoom-in canvas used UMAP with five neighbors and 0.1 minimum distance; when labels were available, the semi-supervised trade-off parameter was fixed at 0.5, penalizing intra-class and global consistencies equally.

\subsection{Study of Parts}
\label{subsec:ablation}

Our approach depends on two main independent steps: the image partition into segments and the interactive labeling of those regions. The inaccuracy of one of them would significantly deteriorate the performance of the feature space annotation for
image segmentation labeling.
Therefore, we present a study of parts that considers two ideal scenarios: (a) interactive projection labeling of perfect segments and (b) image partition into segments followed by optimal labeling. 

In (a), the user annotates segments from a perfect image partition inside and outside the objects' masks. Hence, every segment will always belong to a single class. Table~\ref{tab:annot} reports the results. The Intersection over Union (IoU) distribution is heavily right-skewed, as noticed from the differences between average IoU and median IoU, indicating that most segments were labeled correctly. Visual inspection revealed that user annotation errors occurred only in small components. 
Table~\ref{tab:annot} also reports the total time (in seconds) spent annotating (user) and processing (machine), starting from the initial segment projection presented to the user. It indicates that feature space projection annotation with metric learning is effective for image segmentation annotation.

\begin{table}
\centering
\begin{tabular}{@{}lcccc@{}} \toprule
Dataset & Avg. IoU   & Median IoU & Time (s) \\ \midrule
iCoSeg  & 95.07      & 99.96      & 5.32     \\
DAVIS   & 98.54      & 99.97      & 7.82     \\
Rooftop & 95.10      & 99.99      & 3.96     \\ \bottomrule
\end{tabular}
\caption{Average IoU, median IoU, average total processing time in seconds per image.}
\label{tab:annot}
\end{table}

In (b), we measure the IoU of the watershed hierarchical cut using a fixed parameter --- the threshold of 1000 with the volume criterion. The segments were then labeled by majority vote among the true labels of their pixels. Table~\ref{tab:segm} shows the quality of segmentation, which imposes the upper bound to the quality of the overall projection labeling procedure if no segment correction was performed.

\begin{table}
\centering
\begin{tabular}{@{}lcc@{}} \toprule
Dataset & Avg. IoU   & Median IoU \\ \midrule
iCoSeg  & 84.15      & 91.86      \\
DAVIS   & 82.46      & 88.50      \\
Rooftop & 75.14      & 76.77      \\ \bottomrule
\end{tabular}
\caption{Automatic segmentation results with their respective dataset.}
\label{tab:segm}
\end{table}

For reference, Click Carving~\cite{Jain:2019:ClickCarving}  reports an average IoU of $84.31$ in the iCoSeg, dataset when selecting the optimal segment (\ie highest IoU among proposals) from a pool of approximately 2000 segment proposals per image, produced with MCG~\cite{Pont:2016:MCG}. In contrast, we obtain an equivalent performance of $84.15$ with a fixed segmentation with disjoints candidates only.

\begin{figure}[!ht]
    \centering
    \begin{subfigure}[b]{0.3\columnwidth}
        \centering
        \includegraphics[width=\textwidth]{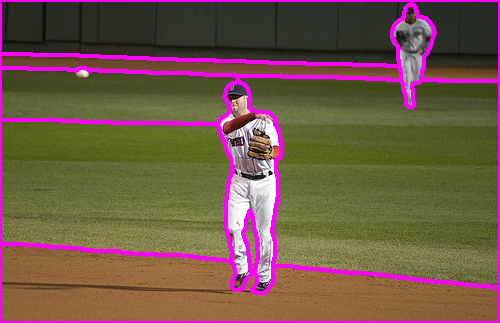}
        \caption{}
    \end{subfigure}
    \begin{subfigure}[b]{0.3\columnwidth}
        \centering
        \includegraphics[width=\textwidth]{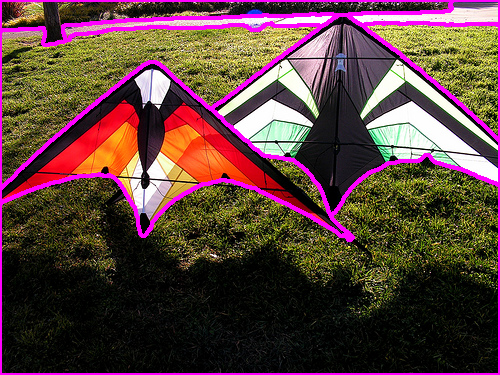}
        \caption{}
    \end{subfigure}
    \begin{subfigure}[b]{0.3\columnwidth}
        \centering
        \includegraphics[width=\textwidth]{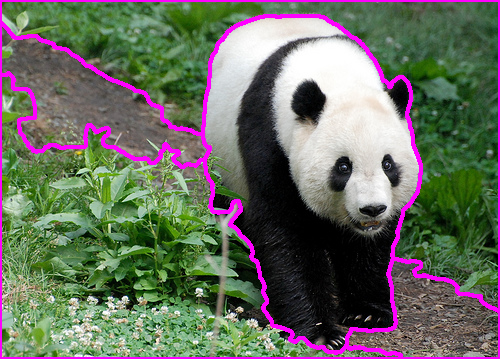}
        \caption{}
    \end{subfigure} \\
    \begin{subfigure}[b]{0.3\columnwidth}
        \centering
        \includegraphics[width=\textwidth]{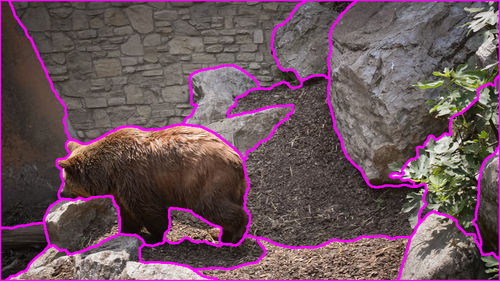}
        \caption{}
    \end{subfigure}
    \begin{subfigure}[b]{0.3\columnwidth}
        \centering
        \includegraphics[width=\textwidth]{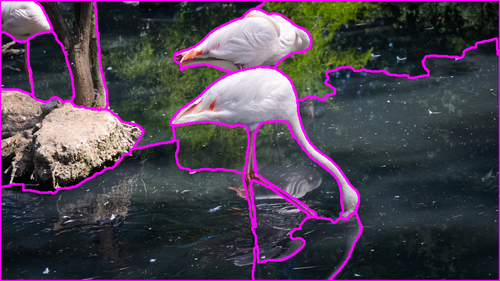}
        \caption{}
    \end{subfigure}
    \begin{subfigure}[b]{0.3\columnwidth}
        \centering
        \includegraphics[width=\textwidth]{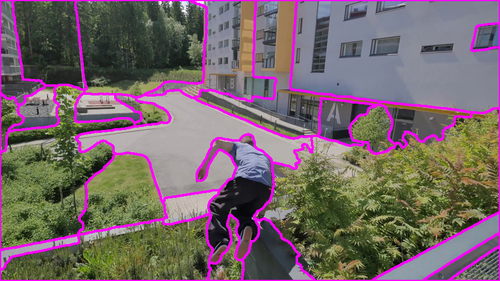}
        \caption{}
    \end{subfigure} \\
    \begin{subfigure}[b]{0.3\columnwidth}
        \centering
        \includegraphics[width=\textwidth]{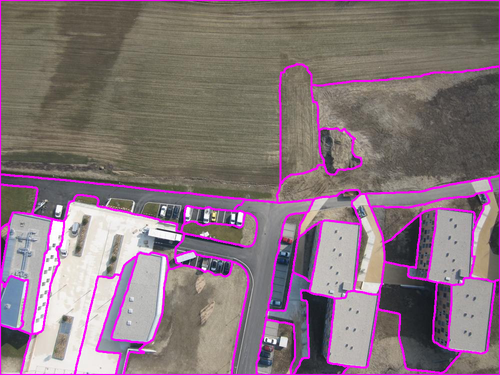}
        \caption{}
    \end{subfigure}
    \begin{subfigure}[b]{0.3\columnwidth}
        \centering
        \includegraphics[width=\textwidth]{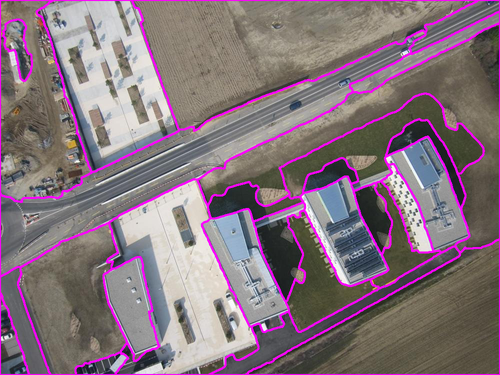}
        \caption{}
    \end{subfigure}
    \begin{subfigure}[b]{0.3\columnwidth}
        \centering
        \includegraphics[width=\textwidth]{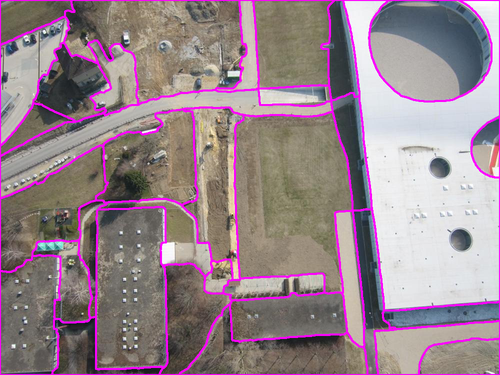}
        \caption{}
    \end{subfigure}
    \caption[Candidate segments from ablation study]{Candidate segments from the study of parts. First row is from the iCoSeg dataset, second is from DAVIS and the last from Rooftop.}
    \label{fig:ablat}
\end{figure}

Figure~\ref{fig:ablat} presents example of the candidate segments on the three datasets.

\subsection{Quantitative analysis using baselines}
\label{subsec:quant}

\begin{table*}
\centering
\begin{tabular}{@{}lccccccccc@{}} \toprule
Dataset           && \multicolumn{2}{c}{iCoSeg}   && \multicolumn{2}{c}{DAVIS}    && \multicolumn{2}{c}{Rooftop}  \\ 
\cmidrule{1-1} \cmidrule{3-4} \cmidrule{6-7} \cmidrule{9-10}
Method            && IoU               & Time (s) && IoU               & Time (s) && IoU               & Time (s) \\ \midrule
f-BRS (3 clicks)  && 79.82             & 4.2      && 79.87             & 4.2      && 62.57             & 4.2      \\
f-BRS (5 clicks)  && 82.14             & 6        && 82.44             & 6        && 74.53             & 6        \\
FCANet (3 clicks) && \underline{84.63} & 4.2      && 82.44             & 4.2      && 65.99             & 4.2      \\
FCANet (5 clicks) && \textbf{88.00}    & 6        && \textbf{86.63}    & 6        && \textbf{81.38}    & 6        \\
Ours              && 84.29             & 5.96     && \underline{84.53} & 8.74     && \underline{77.28} & 7.02     \\ \bottomrule
\end{tabular}
\caption{Average IoU and time over images, except for Rooftop, where time is computed over instances. For robot user experiments, with multiple budgets (3 and 5 clicks), time was estimated according to this study~\cite{Bearman:2016:WhatIsThePoint}. Our method obtains comparable accuracy, but it spends additional time annotating foreground and background. The Cityscapes experiment shows our results on a more realistic scenario where every pixel is labeled (not a microtask).}
\label{tab:quant}
\end{table*}

Since existing baselines report scores only in a very limited scenario, we executed our own experiments according to the code availability; Them being, f-BRS-B~\cite{Sofiiuk:2020:fBRS} and FCANet~\cite{Lin:2020:FirstClickAtte}, both with \textit{Resnet101} backbone, with their publicly available weights~\fakecite{Code:2020:FBRS, Code:2020:FCANet} trained on the SBD~\cite{Hariharan:2011:SemanticContours} and SBD plus PASCAL VOC~\cite{Everingham:2010:PascalVOC} datasets, respectively. We are not comparing with IOG~\cite{Zhang:2020:IOG} because we could not reproduce the results (subpar performance) with their available code and weights, and~\cite{Kontogianni:2020:ContinuousAdapt} is not publicly available.
The results are reported over the final segmentation mask, given a sequence of 3 and 5 clicks.

Table~\ref{tab:quant} report the average IoU and the total time spent in annotation. For click-based methods, the interaction time was estimated as 2.4s for the initial click and 0.9s for additional clicks~\cite{Bearman:2016:WhatIsThePoint}.

\begin{figure*}
    \centering
    \rotatebox{90}{\quad \quad \quad Ground-truth}
    \hfill
    \begin{subfigure}[b]{0.3\textwidth}
        \centering
        \includegraphics[width=\textwidth]{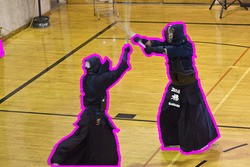}
        \caption{}
    \end{subfigure}
    \hfill
    \begin{subfigure}[b]{0.3\textwidth}
        \centering
        \includegraphics[width=\textwidth]{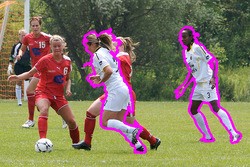}
        \caption{}
    \end{subfigure}
    \hfill
    \begin{subfigure}[b]{0.3\textwidth}
        \centering
        \includegraphics[width=\textwidth]{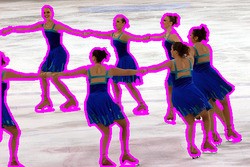}
        \caption{}
    \end{subfigure} \\
    \rotatebox{90}{\quad \quad \quad \quad Ours}
    \hfill
    \begin{subfigure}[b]{0.3\textwidth}
        \centering
        \includegraphics[width=\textwidth]{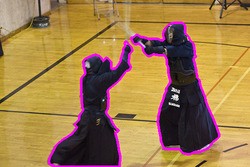}
        \caption{}
    \end{subfigure}
    \hfill
    \begin{subfigure}[b]{0.3\textwidth}
        \centering
        \includegraphics[width=\textwidth]{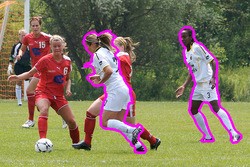}
        \caption{}
    \end{subfigure}
    \hfill
    \begin{subfigure}[b]{0.3\textwidth}
        \centering
        \includegraphics[width=\textwidth]{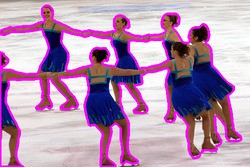}
        \caption{}
    \end{subfigure} \\
        \rotatebox{90}{\quad \quad \quad \quad FCANet}
    \hfill
    \begin{subfigure}[b]{0.3\textwidth}
        \centering
        \includegraphics[width=\textwidth]{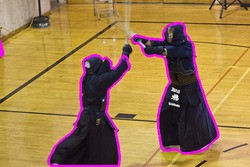}
        \caption{}
    \end{subfigure}
    \hfill
    \begin{subfigure}[b]{0.3\textwidth}
        \centering
        \includegraphics[width=\textwidth]{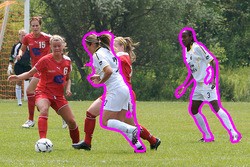}
        \caption{}
    \end{subfigure}
    \hfill
    \begin{subfigure}[b]{0.3\textwidth}
        \centering
        \includegraphics[width=\textwidth]{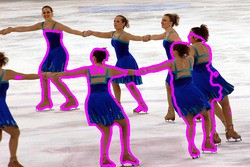}
        \caption{}
    \end{subfigure} \\
    \rotatebox{90}{\quad \quad \quad \quad f-BRS}
    \hfill
    \begin{subfigure}[b]{0.3\textwidth}
        \centering
        \includegraphics[width=\textwidth]{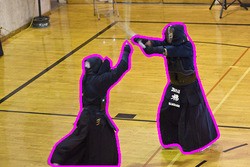}
        \caption{}
    \end{subfigure}
    \hfill
    \begin{subfigure}[b]{0.3\textwidth}
        \centering
        \includegraphics[width=\textwidth]{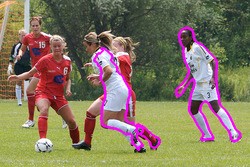}
        \caption{}
    \end{subfigure}
    \hfill
    \begin{subfigure}[b]{0.3\textwidth}
        \centering
        \includegraphics[width=\textwidth]{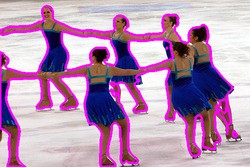}
        \caption{}
    \end{subfigure} \\
    \caption[iCoSeg qualitative results]{The magenta boundaries delineate regions with foreground labels for the ground-truth data, our method, and the baselines using 5 clicks per image on the iCoSeg dataset. Figure (i) shows that FCANet has difficulties when segmenting multiples instances, as mentioned in their original article.}
    \label{fig:icoseg}
\end{figure*}

\begin{figure*}
    \centering
    \rotatebox{90}{\quad \quad \quad Ground-truth}
    \hfill
    \begin{subfigure}[b]{0.3\textwidth}
        \centering
        \includegraphics[width=\textwidth]{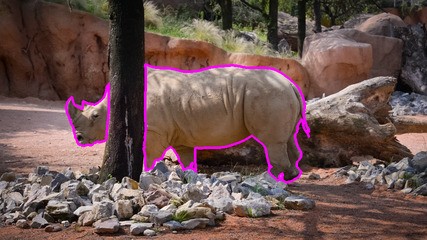}
        \caption{}
    \end{subfigure}
    \hfill
    \begin{subfigure}[b]{0.3\textwidth}
        \centering
        \includegraphics[width=\textwidth]{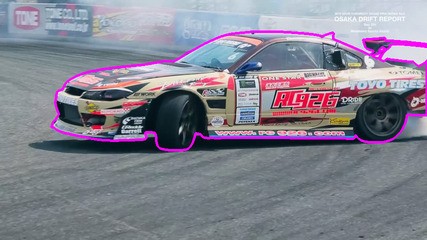}
        \caption{}
    \end{subfigure}
    \hfill
    \begin{subfigure}[b]{0.3\textwidth}
        \centering
        \includegraphics[width=\textwidth]{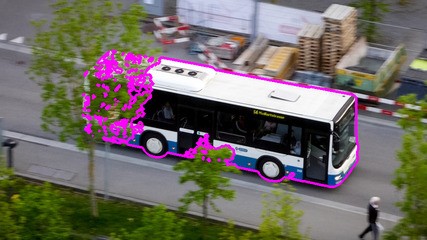}
        \caption{}
    \end{subfigure} \\
    \rotatebox{90}{\quad \quad \quad \quad Ours}
    \hfill
    \begin{subfigure}[b]{0.3\textwidth}
        \centering
        \includegraphics[width=\textwidth]{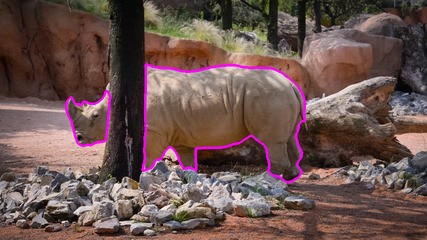}
        \caption{}
    \end{subfigure}
    \hfill
    \begin{subfigure}[b]{0.3\textwidth}
        \centering
        \includegraphics[width=\textwidth]{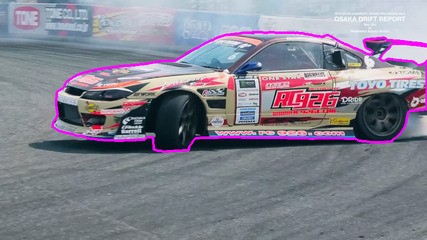}
        \caption{}
    \end{subfigure}
    \hfill
    \begin{subfigure}[b]{0.3\textwidth}
        \centering
        \includegraphics[width=\textwidth]{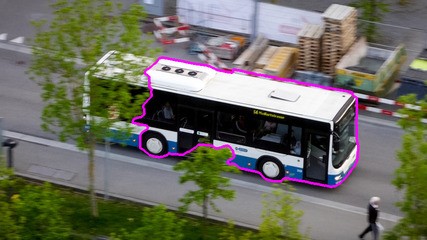}
        \caption{}
    \end{subfigure} \\
     \rotatebox{90}{\quad \quad \quad \quad FCANet}
    \hfill
    \begin{subfigure}[b]{0.3\textwidth}
        \centering
        \includegraphics[width=\textwidth]{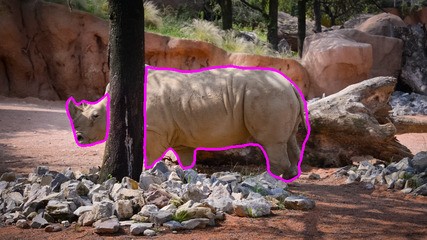}
        \caption{}
    \end{subfigure}
    \hfill
    \begin{subfigure}[b]{0.3\textwidth}
        \centering
        \includegraphics[width=\textwidth]{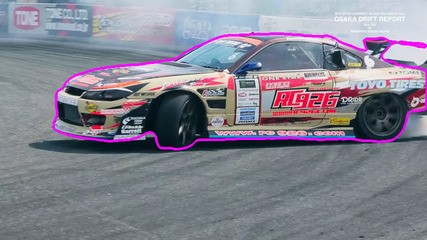}
        \caption{}
    \end{subfigure}
    \hfill
    \begin{subfigure}[b]{0.3\textwidth}
        \centering
        \includegraphics[width=\textwidth]{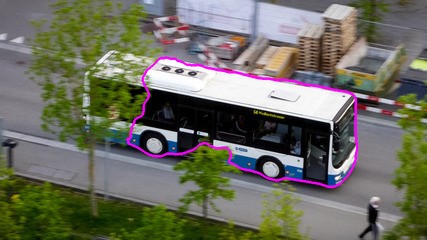}
        \caption{}
    \end{subfigure} \\
    \rotatebox{90}{\quad \quad \quad \quad f-BRS}
    \hfill
    \begin{subfigure}[b]{0.3\textwidth}
        \centering
        \includegraphics[width=\textwidth]{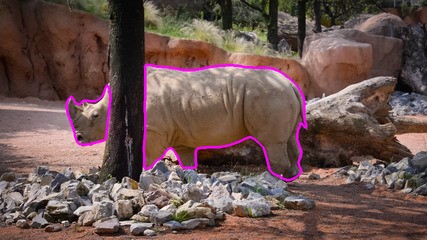}
        \caption{}
    \end{subfigure}
    \hfill
    \begin{subfigure}[b]{0.3\textwidth}
        \centering
        \includegraphics[width=\textwidth]{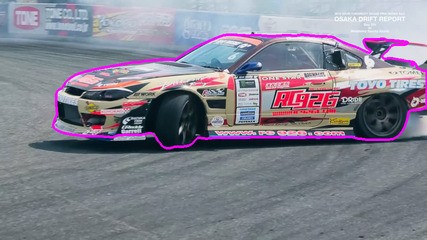}
        \caption{}
    \end{subfigure}
    \hfill
    \begin{subfigure}[b]{0.3\textwidth}
        \centering
        \includegraphics[width=\textwidth]{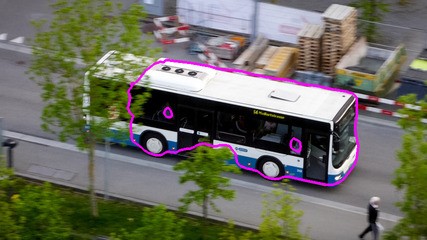}
        \caption{}
    \end{subfigure} \\
    \caption[DAVIS qualitative results]{The magenta boundaries delineate regions with foreground labels for the ground-truth data, our method, and the baselines using 5 clicks per image on the DAVIS dataset.}
    \label{fig:davis}
\end{figure*}

\begin{figure*}
    \centering
    \rotatebox{90}{\quad \quad \quad \quad Ground-truth}
    \hfill
    \begin{subfigure}[b]{0.3\textwidth}
        \centering
        \includegraphics[width=\textwidth]{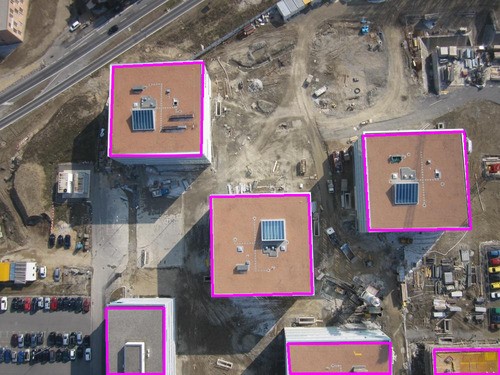}
        \caption{}
    \end{subfigure}
    \hfill
    \begin{subfigure}[b]{0.3\textwidth}
        \centering
        \includegraphics[width=\textwidth]{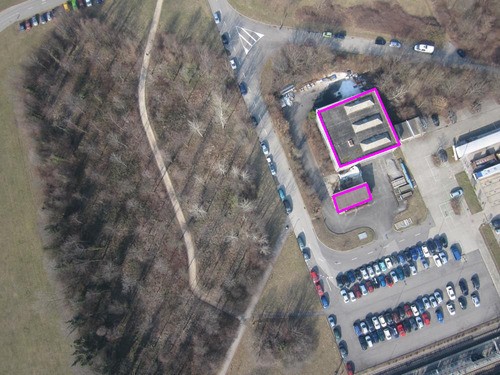}
        \caption{}
    \end{subfigure}
    \hfill
    \begin{subfigure}[b]{0.3\textwidth}
        \centering
        \includegraphics[width=\textwidth]{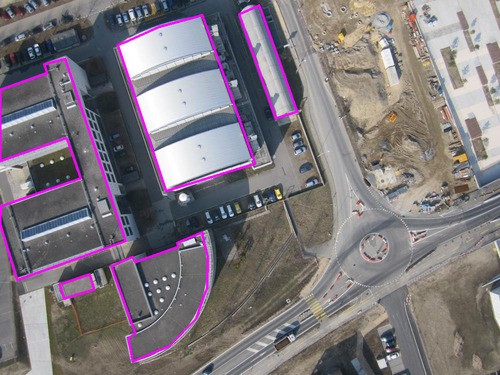}
        \caption{}
    \end{subfigure} \\
    \rotatebox{90}{\quad \quad \quad \quad \quad Ours}
    \hfill
    \begin{subfigure}[b]{0.3\textwidth}
        \centering
        \includegraphics[width=\textwidth]{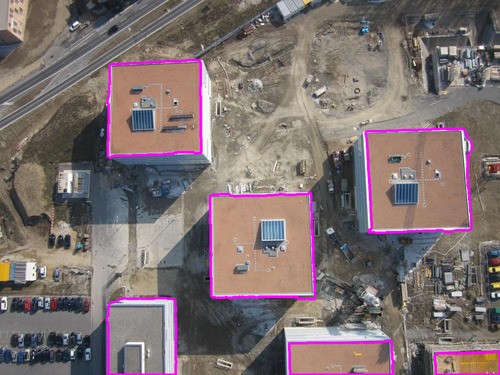}
        \caption{}
    \end{subfigure}
    \hfill
    \begin{subfigure}[b]{0.3\textwidth}
        \centering
        \includegraphics[width=\textwidth]{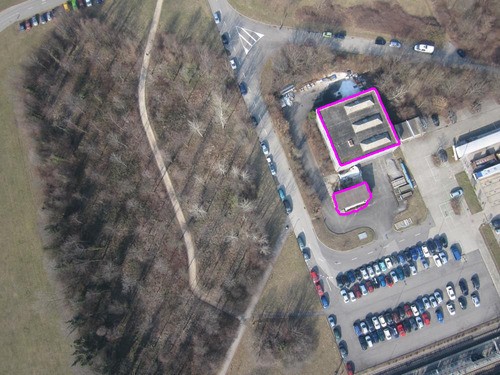}
        \caption{}
    \end{subfigure}
    \hfill
    \begin{subfigure}[b]{0.3\textwidth}
        \centering
        \includegraphics[width=\textwidth]{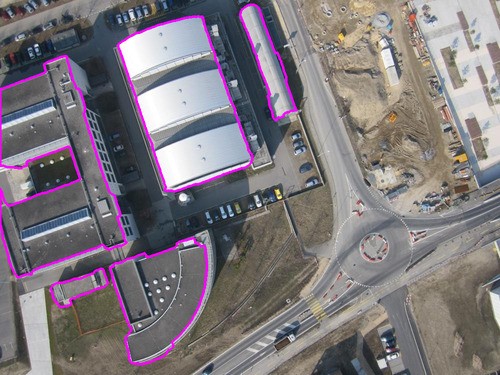}
        \caption{}
    \end{subfigure} \\
    \rotatebox{90}{\quad \quad \quad \quad \quad FCANet}
    \hfill
    \begin{subfigure}[b]{0.3\textwidth}
        \centering
        \includegraphics[width=\textwidth]{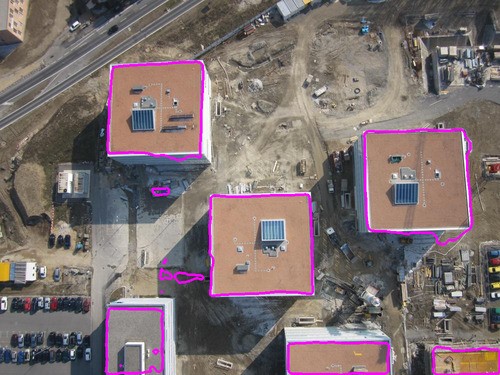}
        \caption{}
    \end{subfigure}
    \hfill
    \begin{subfigure}[b]{0.3\textwidth}
        \centering
        \includegraphics[width=\textwidth]{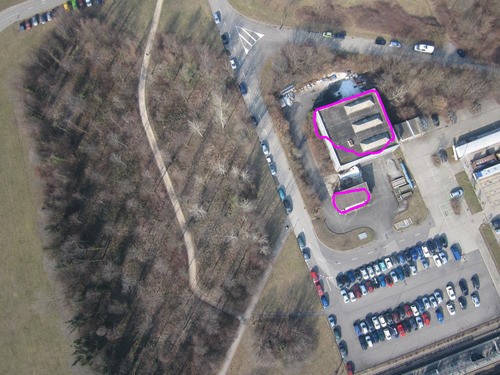}
        \caption{}
    \end{subfigure}
    \hfill
    \begin{subfigure}[b]{0.3\textwidth}
        \centering
        \includegraphics[width=\textwidth]{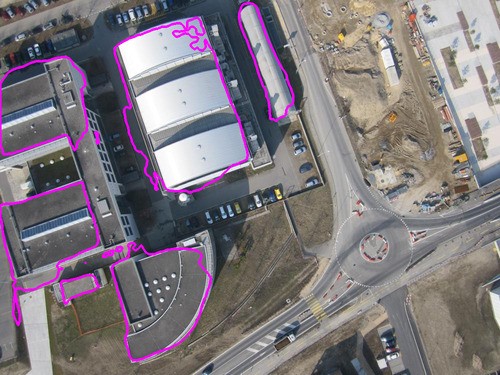}
        \caption{}
    \end{subfigure} \\
    \rotatebox{90}{\quad \quad \quad \quad \quad f-BRS}
    \hfill
    \begin{subfigure}[b]{0.3\textwidth}
        \centering
        \includegraphics[width=\textwidth]{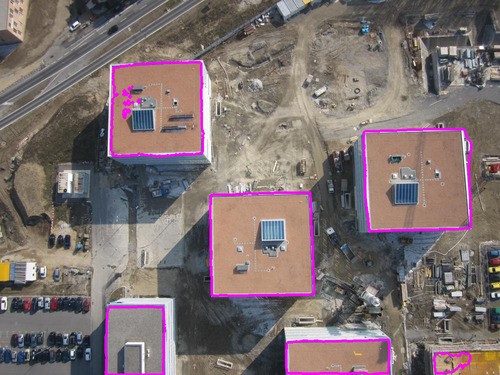}
        \caption{}
    \end{subfigure}
    \hfill
    \begin{subfigure}[b]{0.3\textwidth}
        \centering
        \includegraphics[width=\textwidth]{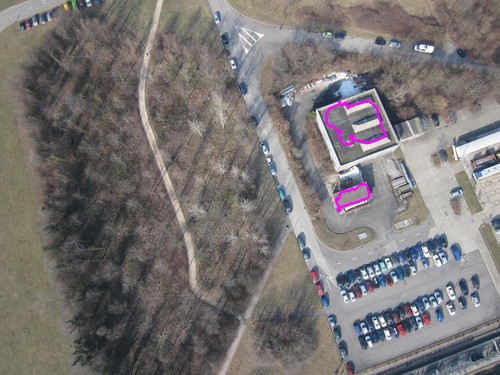}
        \caption{}
    \end{subfigure}
    \hfill
    \begin{subfigure}[b]{0.3\textwidth}
        \centering
        \includegraphics[width=\textwidth]{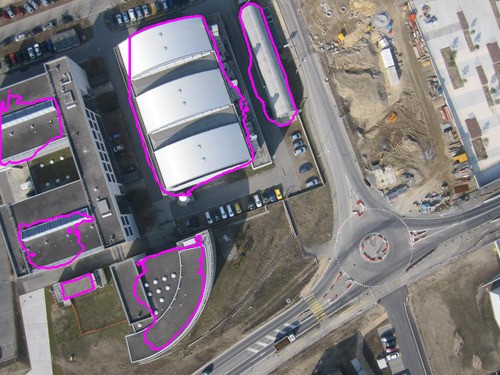}
        \caption{}
    \end{subfigure} \\
    \caption[Rooftop qualitative results]{The magenta boundaries delineate regions with foreground labels for the ground-truth data, our method, and the baselines using 5 clicks per instance on the Rooftop dataset.}
    \label{fig:rooftop}
\end{figure*}

We achieve comparable accuracy results with state-of-the-art methods while employing less sophisticated segmentation procedures, qualitative results are presented in Figures~\ref{fig:icoseg}-~\ref{fig:rooftop}. Despite this, existing methods require less time to annotate these datasets; this is due to them being specialized in the foreground annotation microtask, while our approach wastes time annotating the background --- this is exacerbated on the DAVIS dataset where a background object might be a category equal to the foreground.

The following experiment evaluates our performance on a semantic segmentation, where labeling the whole image is the final goal, not just the microtask of delineating a single object.

\subsection{Qualitative analysis for semantic segmentation}
\label{subsec:qual}

To verify the proposed approach in a domain-specific scenario, we evaluate its performance on Cityscapes~\cite{Cordts:2016:Cityscapes}. Since the true labels of the test set are not available, we took the same approach as ~\cite{Castrejon:2017:PolyRNN}, by testing on the validation set. Further, the annotation quality was evaluated on 98 randomly chosen images (about 20\% of the validation set). 

PoolNet was optimized based on the boundaries of the training set's true labels for five epochs using the Adam optimizer, a learning rate of $5e^{-5}$, weight decay of $5e^{-4}$, and a batch of size 8. Predictions were performed on a single scale. The fine-tuned gradient ignores irrelevant boundaries, reducing over segmentation.

\begin{figure}
    \centering
    \rotatebox{90}{\quad \quad Image}
    \hfill
    \begin{subfigure}[b]{0.45\columnwidth}
        \centering
        \includegraphics[width=\textwidth]{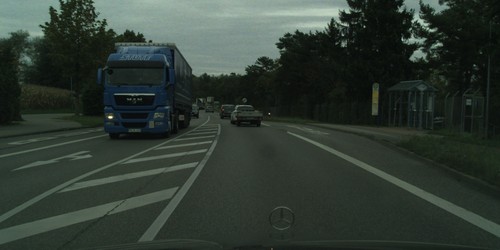}
    \end{subfigure}
    \hfill
    \begin{subfigure}[b]{0.45\columnwidth}
        \centering
        \includegraphics[width=\textwidth]{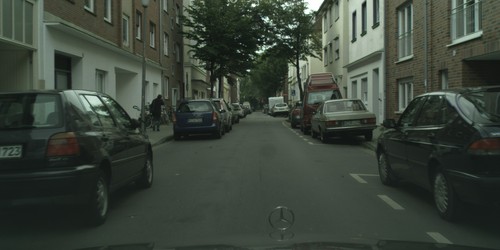}
    \end{subfigure}
    \hfill \\
    \rotatebox{90}{\quad Ground-truth}
    \hfill
    \begin{subfigure}[b]{0.45\columnwidth}
        \centering
        \includegraphics[width=\textwidth]{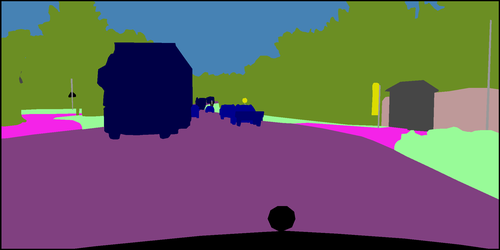}
    \end{subfigure}
    \hfill
   \begin{subfigure}[b]{0.45\columnwidth}
        \centering
        \includegraphics[width=\textwidth]{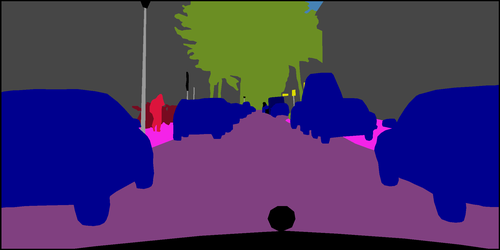}
    \end{subfigure}
    \hfill \\
    \rotatebox{90}{Our Results}
    \hfill
    \begin{subfigure}[b]{0.45\columnwidth}
        \centering
        \includegraphics[width=\textwidth]{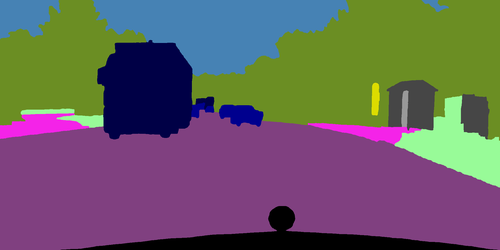}
    \end{subfigure}
    \hfill
    \begin{subfigure}[b]{0.45\columnwidth}
        \centering
        \includegraphics[width=\textwidth]{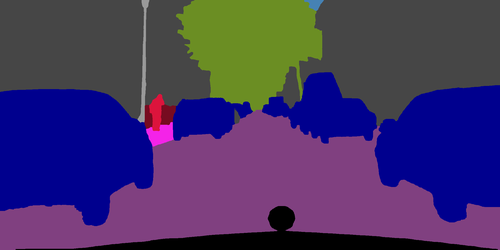}
    \end{subfigure}
    \hfill \\
    \caption{Cityscapes result, each column is a different image, row indicates which kind.}
    \label{fig:cs}
\end{figure}

The original article reports an agreement (\ie accuracy) between annotators of 96\%. We obtained an agreement of 91.5\% with the true labels of the validation set (Figure~\ref{fig:cs}), while spending less than 1.5\% of their time --- \ie our experiment took 1 hour and 58 minutes to annotate the 98 images, while to produce the same amount of ground-truth data took approximately 6.1 days (average of 1.5 hour per image~\cite{Cordts:2016:Cityscapes}) --- about 74.75 times faster than the original procedure.
These 98 images contain in total 6500 default classes' polygons (\ie, instances). Thus, with the estimate of 6 secs per instance, FCANet would take 10 hours and 50 minutes to label them.

\section{Conclusion}
\label{sec:discussion}

We presented a novel interactive image segmentation paradigm for simultaneous annotation of segments from multiple images in their deep features' projection space. Despite employing less sophisticated segmentation methods, it achieves comparable performance with more modern approaches.
Moreover, \red{existing interactive segmentation methodologies are not direct competitors and can complement the presented implementation, for example FCANet could be used for segmentation correction and Parametric UMAP~\cite{Sainburg:2020:ParamUMAP} to unify the embedding and feature extraction in an end-to-end procedure.}

We believe that this work leads to several opportunities for combining the whole pipeline into a holistic segmentation procedure, where redundant samples are labeled at once, and annotation on the image domain occurs only when necessary.

In a real scenario, \red{to decrease the users' cognitive load and to annotate in a distributed manner,} we suggest letting a leading user interact with the projection to evaluate existing annotated data, model performance (segments clustering), and, when necessary, delegating segment correction to workers, as it is currently done in existing annotation procedures, diminishing the total images annotated on the image domain. 

\red{The proposed approach can also benefit applications beyond the labeling of standard computer vision datasets. For example, in biology, spatial transcriptomics~\cite{Moses:2022:TranscriptomicsMuseum} datasets contain segments (\ie cells) and their features (\ie transcriptomes) which naturally fall into our framework, therefore, they could be explored and labeled using the proposed pipeline.}

\section*{Acknowledgements}

\noindent
This work was supported by CAPES-COFECUB [\#42067SF]; CNPq [\#303808/2018-7]; and FAPESP research grants [\#2014/12236-1, \#2019/11349-0, and \#2019/21734-9];

\bibliography{references}

\end{document}